%% file: main.tex
\documentclass[11pt,letterpaper]{article}
\usepackage[margin=0.9in]{geometry}
\usepackage[T1]{fontenc}
\usepackage[utf8]{inputenc}
\usepackage{lmodern}
\usepackage{amsmath,amssymb,booktabs,array,tabularx,graphicx}
\usepackage[numbers,sort&compress]{natbib}
\usepackage{microtype,xcolor,xurl,enumitem,tikz}
\usetikzlibrary{positioning,arrows.meta}
\usepackage{hyperref}
\usepackage[all]{hypcap}
\usepackage[section]{placeins}
\usepackage{needspace}
\definecolor{ink}{HTML}{19354A}
\definecolor{wash}{HTML}{F0F4F6}
\hypersetup{hidelinks,pdftitle={DFAH-Bench: What Agreement Leaves Unmeasured in Financial Agents},pdfauthor={Raffi Khatchadourian},pdfsubject={Replay agreement, record completeness, and evidence for actions in financial agents}}
\setlist{itemsep=3pt,topsep=5pt,leftmargin=*}
\newcommand{\dar}{\mathrm{DAR}}

\newcommand{\tarseq}{\mathrm{TAR}_{\mathrm{seq}}}

\newcommand{\tarstrong}{\mathrm{TAR}_{\mathrm{strong}}}
\newcommand{\gapmetric}{\Delta_{\mathrm{DT}}}
\newcommand{\pp}{\,\mathrm{pp}}
\newcommand{\allow}{\texttt{allow}}
\newcommand{\block}{\texttt{block}}
\newcommand{\review}{\texttt{review}}
\providecommand{\Description}[1]{}
\newcolumntype{Y}{>{\raggedright\arraybackslash}X}
\title{\textbf{DFAH-Bench: What Agreement Leaves\\Unmeasured in Financial Agents}}
\author{Raffi Khatchadourian\\IBM Financial Services Market, New York, NY, USA\\\small\texttt{raffi.khatchadourian1@ibm.com}}
\date{}
\begin{document}
\maketitle
\begin{abstract}
When a financial agent repeats its decision, how much does that tell us about the actions behind it? DFAH-Bench examines this question through replay experiments, a controlled test of omitted tool calls, and banking tasks with authorization checks. Across 570 eligible prospective replay episodes, decisions agreed at rates of \mbox{94--95\%}, while tool paths, including inputs and results, agreed at \mbox{45--52\%}; one condition fell one group short of the planned minimum. In a separate 48-execution diagnostic, deleting the same calls from repeated records left agreement perfect. In banking probes, repeated authorization decisions sometimes conflicted with constructed rule labels. The $\tau$-Knowledge banking study scheduled 1,080 episodes, with 1,033 known outcomes. Structural checks alone completed more primary tasks than either model-based gate-and-recovery bundle. Input limits and recovery behavior materially shape these results. The typed-choice bundle had lower mean episode cost than the generative bundle on complete task pairs, but fewer successes across the primary schedule under every assignment of unknown outcomes. Missing outcomes prevented the planned confirmatory tests, so the banking comparisons are descriptive. Together, the findings show that agreement alone cannot establish reliability: it can conceal changes in execution, systematic omissions, and unsupported decisions. Evaluation therefore needs to connect what the agent did, the evidence and rules governing its action, and whether it completed the task.
\end{abstract}
\noindent{\small\textbf{Code and data:} \url{https://github.com/ibm-client-engineering/output-drift-financial-llms}}\par

\input{introduction}
\input{replay}
\input{cancellation}
\input{capture}
\input{banking}
\input{discussion}
\FloatBarrier
\clearpage
\bibliographystyle{plainnat}
\bibliography{references}
\clearpage
\appendix

\input{replay_appendix}

\input{capture_appendix}
\input{banking_appendix}
\input{provenance}

\end{document}

%% file: introduction.tex
\section{Agreement as evidence about agent behavior}
\label{sec:introduction}

A financial agent may reach the same final decision through different checks. It may also repeat a decision because it repeatedly overlooks the same fact. An evaluator that sees only the final answer cannot distinguish these situations. Recording the path helps, but then a second question arises: does that record preserve the actions that actually ran? If a recorded judgment is used to permit an action, a third question follows: did the available evidence justify that permission?

The common question is what agreement leaves unmeasured. The Determinism--Faithfulness Assurance Harness (DFAH) provides the measurement harness; DFAH-Bench is the associated benchmark and experimental record. In the harness name, \emph{faithfulness} refers to replay fidelity of the recorded execution path. Capture fidelity separately asks whether that record preserves what actually ran. Decision agreement, path agreement, capture fidelity, policy-label match, and task completion each require their own observations.

We contribute an empirical account of three measurement limits. Replay analysis measures changes in recorded behavior behind repeated decisions. Retained-case and controlled-deletion examples show how averaging or missing observations can hide such changes. Banking interventions and fixed-state probes separate repeated permission decisions from rule match, decision availability, task completion, and expenditure. These studies belong together because interpreting agreement requires knowing what was recorded, whether the evidence supports the action, and what work followed. Each study tests a separate part of that reasoning.

The studies use separate populations and protocols (Figure~\ref{fig:questions}). Replays compare a fixed case across repeated executions. Banking interventions change which proposed actions execute and can therefore change later dialogue and recovery attempts; those comparisons concern the implemented systems as a whole.

\begin{figure}[htbp]
\centering
\begin{tikzpicture}[font=\small,box/.style={draw=ink!55,fill=wash,rounded corners=2pt,inner sep=8pt,text width=1.87in,minimum height=1.28in,align=left},>=Latex]
\node[box] (repeat) {\textbf{1. Does behavior repeat?}\\[5pt]Compare the final decision and recorded calls across the same case.\\[5pt]\textit{Replay studies and cancellation example}};
\node[box,right=0.13in of repeat] (capture) {\textbf{2. Is the record complete?}\\[5pt]Compare recorded calls with the tested execution reference.\\[5pt]\textit{Paired call-deletion diagnostic}};
\node[box,right=0.13in of capture] (permit) {\textbf{3. Is action justified?}\\[5pt]Inspect prerequisites, permission, execution, and outcomes.\\[5pt]\textit{Banking interventions and gate probes}};
\end{tikzpicture}
\caption{Three connected questions with separate tests. Agreement can be high even when a record is incomplete or a prerequisite is missing. The tests retain their own cases, denominators, and limits.}
\label{fig:questions}
\end{figure}

\subsection{Study design and evidence}
Table~\ref{tab:evidence-map} identifies each study's unit and question.

This paper connects the banking experiments and gate probes described in the AI4F manuscript \citep{khatchadourian2026ai4f} with the local cancellation and capture studies described in the SLM-Agents manuscript \citep{khatchadourian2026slmagents}. Together with the broader replay studies, they examine what agreement leaves unmeasured. Shared observations are counted once: the 800 Qwen episodes, for example, are part of the 4,157 historical episodes. Appendix~\ref{app:provenance} identifies the source studies and available research materials.

\begin{table}[htbp]
\centering\small
\caption{Evidence used in this paper. Counts describe different units and must not be added as if they were independent tasks. ``Eligible'' means the required observations were available under that study's rules.}
\label{tab:evidence-map}
\begin{tabularx}{\linewidth}{@{}>{\raggedright\arraybackslash}p{0.25\linewidth}>{\raggedright\arraybackslash}p{0.29\linewidth}Y@{}}
\toprule
Study & Retained scale & Question answered \\
\midrule
Historical replays & 4,157 episodes; 719 case groups; eight configurations & Does a repeated decision conceal a changed tool-name path? \\
Prospective API replays & 570/600 episodes; 190/200 groups & What changes when arguments and results are also recorded? \\
Local reconciliation check & 792/800 episodes; 99/100 groups & Does a fixed four-tool workflow repeat within its capture boundary? \\
Capture diagnostic & 48 executions; 16 groups; 144 paired record views & Can systematic omissions preserve perfect agreement? \\
Interactive banking & 1,080 episodes; 1,033 known native outcomes & How do implemented gates affect completion, availability, and API cost? \\
Fixed-state gate probes & 648 constructed-rule; 1,056 representation; 240 selected-case requests & Do verdicts match constructed rules or change with visible evidence? \\
\bottomrule
\end{tabularx}
\end{table}

\section{What agreement measures}
\label{sec:metrics}

A \emph{case} is one task with fixed input and fixture data. A \emph{configuration} includes the model, provider, sampling controls, tools, and harness. A \emph{replay group} contains repeated episodes for the same case and configuration. Holding those recorded settings fixed does not establish that every provider-side computation is identical.

Decision agreement (DAR) is the fraction of repeats giving the most common final decision. Tool-path agreement (TAR) is the fraction following the most common recorded path. For example, if all three repeats return \emph{escalate}, DAR is 1. If each has a different path, TAR is $1/3$. The most common decision and path are found separately; they need not occur together on a particular repeat.

For eligible case $c$, with $N_c$ repeats, decisions $d_{cr}$ and paths $t_{cr}$,
\begin{align}
\dar_c &= \frac{\max_y\sum_{r=1}^{N_c}\mathbf{1}[d_{cr}=y]}{N_c}, &
\mathrm{TAR}_{\mathrm{seq},c} &= \frac{\max_t\sum_{r=1}^{N_c}\mathbf{1}[t_{cr}=t]}{N_c},\\
\Delta_{\mathrm{DT},c} &= \dar_c-\mathrm{TAR}_{\mathrm{seq},c}. &&
\end{align}
Here $\mathbf{1}$ counts repeats satisfying the condition. The gap compares the two agreement rates within the case. Positive values mean that decisions agree more often than paths; negative values mean the reverse. Averaging the gaps can hide opposing changes, as Section~\ref{sec:cancellation} demonstrates. A more direct inspection trigger is a unanimous decision with more than one path. We report its frequency among unanimous-decision groups and distinguish that denominator from the share of all cases flagged for review.

\paragraph{What counts as the same path?}
Historical records support only the ordered sequence of tool names. Prospective records also retain arguments and results. For those records, a strong path is the ordered sequence
\[
S_{cr}=\big((n_1,a_1,h(v_1)),\ldots,(n_m,a_m,h(v_m))\big),
\qquad
\mathrm{TAR}_{\mathrm{strong},c}=\frac{\max_s\sum_r\mathbf{1}[S_{cr}=s]}{N_c},
\]
where $n_j$ is the tool name, $a_j$ its canonical arguments, and $h(v_j)$ a hash of its canonical returned value. Canonicalization removes irrelevant JSON formatting while preserving types and array order. The September capture collector also retains the returned value itself. The exact byte encodings differ by cohort and are preserved in Appendix~\ref{app:contract}; hashes are not compared across cohorts.

Adding arguments makes the comparison more specific: two calls with the same name may operate on different inputs. Finer representations can only preserve or reduce exact agreement, so the empirical question is the size and content of the additional variation. A decision label also occupies a much smaller outcome space than a full path. The gap is therefore an inspection signal, not a calibrated error or harm score; judging a difference requires the task's policy and state.

\paragraph{An observed empty path is different from a missing record.}
An empty sequence can mean the agent used no tool. A missing, malformed, or incomplete required channel cannot establish that. Each protocol first determines eligibility, then scores the retained groups using the same episodes for DAR and TAR. The capture diagnostic further checks that a populated trajectory field preserves the dispatched calls. Agreement describes the qualified record; it cannot recover evidence that was never captured.

\paragraph{A worked example from the historical corpus.}
In Gemini 2.0 Flash DataOps case DQ-030, all three decisions were \emph{escalate}:
\begin{center}
\small
\begin{tabularx}{\linewidth}{@{}lYl@{}}
\toprule
Replay & Recorded tools, in order & Decision \\
\midrule
1 & historical fixes $\rightarrow$ reference data $\rightarrow$ human escalation & escalate \\
2 & reference data $\rightarrow$ historical fixes & escalate \\
3 & historical fixes $\rightarrow$ reference data & escalate \\
\bottomrule
\end{tabularx}
\end{center}
DAR is 1 and TAR is $1/3$. Replays 2 and 3 differ in order; replay 1 also contains a different tool set. A reviewer would next check whether the case required human escalation. The agreement scores locate that question but do not answer it.

%% file: replay.tex
\section{Stable decisions can conceal different paths}
\label{sec:replay}

\paragraph{Tasks and case construction.}
The historical and prospective API studies use 50 compliance-triage cases
and 50 financial-DataOps cases (Table~\ref{tab:replay-tasks}). The compliance
agent assesses a transaction alert; the DataOps agent selects a disposition
for a data-quality exception. Cases were programmatically authored to
exercise intended control branches, with fixed prompts, tool schemas,
implementations, and synthetic data. They use no real customer or transaction
records. The separate local reconciliation study uses 50 cases with a fixed
four-tool workflow. Fixture labels define each task's decision vocabulary;
financial correctness has not been independently adjudicated.

\begin{table}[htbp]
\centering\small
\caption{Synthetic replay workflows. Tool descriptions summarize the
available operations; individual compliance and DataOps cases need not use
every tool. The eight-case capture diagnostic is a separate study.}
\label{tab:replay-tasks}
\begin{tabularx}{\linewidth}{@{}>{\raggedright\arraybackslash}p{0.23\linewidth}>{\raggedright\arraybackslash}p{0.31\linewidth}Y@{}}
\toprule
Task & Decision vocabulary & Example tool operations \\
\midrule
Compliance triage & dismiss / investigate / escalate & sanctions checks, customer profiles, risk scoring \\
Financial DataOps & auto-fix / escalate / quarantine & exception details, reference data, historical fixes, validation \\
Local reconciliation & reconcile / investigate / escalate & internal record, external record, policy, variance \\
\bottomrule
\end{tabularx}
\end{table}

\subsection{Historical records: a bounded comparison}
The primary historical slice retains 4,157 episodes in 719 groups across eight configurations, subject to the incomplete prefixes described below. We average case measures within a task and then weight the available tasks equally for each configuration. Appendix~\ref{app:task-results} preserves the full configuration and task tables.

Four local configurations used eight repeats, temperature zero, and seed 42. Gemini API configurations used three repeats at temperature zero without a provider seed. The Claude requests omitted the intended temperature despite zero-valued log metadata; those configurations therefore remain a separate provider-default sensitivity set. Gemini 2.5 Pro covers only a 44-case compliance prefix; Claude Opus 4 covers all 50 compliance cases and a 25-case DataOps prefix. Their missing coverage is unresolved.

Under these protocols, the complete-coverage Gemini 2.0 Flash configuration has 93.7\% decision agreement and 88.7\% ordered tool-name agreement, a 5.0-percentage-point gap. Among its 81 unanimous-decision groups, 20 still have varying paths (24.7\%). The limited Pro prefix has 14 such groups among 30 unanimous groups (46.7\%). The four local configurations have near-zero average gaps, with varying paths in 0--5.9\% of unanimous groups. Complete configurations share the two 50-case tasks and name-only path metric.

The pooled, unweighted historical count is 122 varying paths among 627 unanimous-decision groups. Of the 122, 17 change only tool order, 58 change multiplicity while retaining the distinct tool names, and 47 change the set of names. Among the 47 set-changing groups, 36 vary a control or action contact, 14 an evidence contact, and three both. The records identify the changes; they do not establish which contacts were required in every case.

\paragraph{What the historical data cannot resolve.}
Task coverage, repeat counts, providers, sampling controls, and harnesses differ across the historical configurations, preventing an architecture ranking or attribution of variation to a single component. A legacy defect could also substitute a default decision label after parsing failure without recording that substitution, making some behavioral interpretations of DAR uncertain. Conservative stress bounds and replay-count sensitivities remain in Appendix~\ref{app:sensitivity}. Subsampling the eight-repeat local records to three repeats leaves their aggregate gaps near zero but does not match the remaining conditions. Historical intervals use 5,000 within-task case resamples with equal task weights. They describe exchangeable reweighting of the authored cases, conditional on the observed repeats.

Two configurations with no tool use were excluded from primary path analysis after the run, while observed empty paths within retained configurations stayed eligible. Removing an inconsistent portfolio fixture and an irreversibly damaged evidence-contact measure also narrowed the original corpus. Appendix~\ref{app:revision} records these corrections and the sensitivity to including the no-tool configurations. We use the historical findings as inspectable counterexamples and rely on the separately collected prospective records for argument-aware comparison.

\subsection{Argument-aware comparison reveals additional variation}
On July 23, 2026, GPT-5.6 Terra and Claude Sonnet 5 each ran three replays of the 50 compliance and 50 DataOps cases. Requests used the identifiers \texttt{gpt-5.6-terra} and \texttt{claude-sonnet-5}; successful responses returned those identifiers. The protocol used medium provider-native reasoning, omitted sampling parameters, an eight-turn limit, and at most 2,048 output tokens per turn. Parsing failed closed, and the collector retained names, arguments, and deterministic results.

All 600 scheduled episodes reached terminal states. Comparing a case required three repeats with valid decision and tool-path records. Excluding groups with an invalid required record left 570 episodes in 190 groups. We report this extension separately from the historical primary analysis; Appendix~\ref{app:prospective-coverage} explains the exclusions and the coverage requirement it did not meet.

\begin{table}[htbp]
\centering\small
\caption{Prospective API replay results, task-weighted percentages. Each eligible group has three repeats. Strong TAR compares ordered names, arguments, and result hashes; gaps use unrounded values.}
\label{tab:prospective}
\begin{tabular}{@{}lrrrrr@{}}
\toprule
Configuration & Groups & DAR & Name TAR & Strong TAR & DAR $-$ strong TAR \\
\midrule
GPT-5.6 Terra & 96/100 & 95.1 & 69.4 & 51.5 & 43.6 \\
Claude Sonnet 5 & 94/100 & 94.2 & 66.9 & 45.0 & 49.2 \\
\bottomrule
\end{tabular}
\end{table}

Decision agreement exceeds 94\%, while strong-path agreement is 45.0--51.5\% (Table~\ref{tab:prospective}). Adding canonical arguments reduces agreement from 69.4\% to 51.5\% for Terra and from 66.9\% to 45.0\% for Sonnet. Each comparison uses the same eligible prospective cases and repeats. Because results are deterministic functions of fixture data and arguments here, result hashes introduce no further split once names and arguments match. Appendix~\ref{app:components} retains that decomposition.

Conditioning on unanimous decisions gives the same substantive finding. After equal task weighting, paths still vary in 66.7\% of Terra's unanimous groups and 68.9\% of Sonnet's. These weighted rates do not have a single pooled numerator and denominator without changing the calculation.

\paragraph{A prospective example: the same tools retrieve different history.}
A post hoc inspection of Sonnet DataOps case \texttt{DQ-003} illustrates the
additional information. All three repeats call exception
details, reference data, historical fixes, and human escalation in that order,
then return \emph{escalate}. The historical-fix lookup nevertheless changes:
repeat~1 passes the \texttt{pattern} value
\texttt{missing cusip ticker corporate\_actions} and receives no matching
history. Repeats~2 and~3 pass \texttt{missing\_cusip\_ticker\_lookup} and
receive one synthetic history entry recording 50 earlier cases with an
\emph{escalate} resolution. The \texttt{reason} argument to human escalation
also differs across all three repeats. Thus decision and tool-name agreement
are both 1, while strong-path agreement is $1/3$. This retained example shows
both a lookup value that changes the retrieved evidence and free-text
variation counted by exact path equality. It does not establish a difference
in task correctness or financial risk.

\begin{figure}[htbp]
\centering
\includegraphics[width=0.82\linewidth]{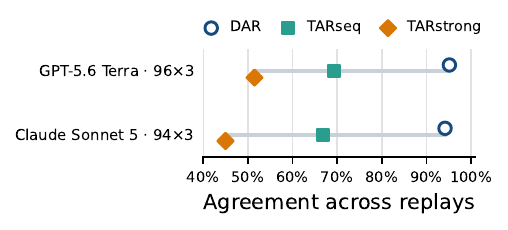}
\caption{The same prospective episodes viewed at three levels: final decision, ordered tool names, and names plus arguments and results. More complete recording reveals variation that an answer-only score misses.}
\label{fig:prospective}
\end{figure}

\paragraph{The separate local reconciliation check.}
Gemma 4 E4B and Qwen 3.5 each completed 400 scheduled episodes on a different 50-case fixture with eight repeats. Eight Qwen formatting failures in one case left 392 episodes in 49 eligible groups; Gemma retained all 400 in 50 groups. No default label was substituted. The decision, final text, and required four-tool argument-and-result path repeated exactly within each eligible group. The same four evidence operations were required in every case. These results establish stability within that fixed workflow and capture boundary.

%% file: cancellation.tex
\subsection{Equal averages can conceal a changing path}
\label{sec:cancellation}

The historical Qwen 2.5 7B configuration gives a concrete counterexample to
reading a zero aggregate gap as absence of path variation. Its retained slice
contains 800 episodes: eight repeats for each of 50 compliance and
50 DataOps cases. These episodes are part of the historical corpus.

Both task-weighted agreement scores equal 99.625\%. Ninety-eight of the
100 groups have zero decision--path gap. Two DataOps cases supply equal,
opposite contributions (Table~\ref{tab:cancellation}). On \texttt{DQ-011},
every decision agrees, while only seven of eight tool-name paths match the
modal path. On \texttt{DQ-040}, seven decisions agree while every path matches.

\begin{table}[htbp]
\centering\small
\caption{The two nonzero case gaps in the retained Qwen 2.5 7B slice.
Fractions use eight repeats per case. The modes for decisions and paths are
chosen independently.}
\label{tab:cancellation}
\begin{tabular}{@{}lrrr@{}}
\toprule
Case & \(\dar\) & \(\tarseq\) & DAR $-$ TAR \\
\midrule
\texttt{DQ-011} & 1.000 & 0.875 & $+0.125$ \\
\texttt{DQ-040} & 0.875 & 1.000 & $-0.125$ \\
\bottomrule
\end{tabular}
\end{table}

\begin{quote}\small
\begin{minipage}{\linewidth}
\textbf{DQ-011, modal recorded path (7/8 repeats):}\\
\texttt{get\_exception\_details}
$\rightarrow$ \texttt{get\_historical\_fixes}
$\rightarrow$ \texttt{query\_reference\_data}\\
$\rightarrow$ \texttt{validate\_fix}
$\rightarrow$ \texttt{apply\_fix}.\\
\textbf{Exceptional path (1/8):} the same first four calls,
ending at \texttt{validate\_fix}. All eight recorded decisions are
\texttt{auto\_fix}.

\textbf{DQ-040:} recorded decisions are \texttt{quarantine} in seven repeats
and \texttt{auto\_fix} in one. All eight recorded paths match the five-call
modal path above.
\end{minipage}
\end{quote}
The display uses normalized historical records. Parser fallbacks cannot be distinguished from model decisions in these records (Section~\ref{sec:replay}).

Because both cases belong to the same task, their gaps cancel within DataOps
and under equal weighting of the two tasks. Yet one of the 97
unanimous-decision groups has a varying path: \mbox{\(1/97\approx1.03\%\)}. The proposed
review trigger would therefore flag one of all 100 eligible groups, a
\(1/100=1\%\) workload before review. The first denominator describes
variation conditional on an unchanged decision; the second describes how
many cases require inspection.

The changed path in \texttt{DQ-011} also changes the distinct tool-name set. Its financial significance needs case review, subject to the historical recording limits in Section~\ref{sec:replay}. The observed result is narrower: averaging the gaps removes an exception that remains visible in the case-level records.

%% file: capture.tex
\section{When the record is incomplete}
\label{sec:capture}

The replay results depend on the record being compared. To test what agreement misses when that record loses information, we deleted calls from recorded paths and compared them with a second record taken at tool completion.

\paragraph{What we ran.}
On September 6, 2026, we froze eight synthetic arithmetic-reconciliation cases,
three repeats per case, two local configurations, tool schemas, prompts, model
manifest digests, and the collection schedule. Four cases require ledger,
statement, and tolerance values from three deterministic, read-only tools.
Four supply all values in the prompt and permit execution without tools.
Qwen 3.5 and Granite 4.1 3B each completed 24 episodes: 48 executions in
16 eligible case--configuration groups. All three planned repeats had to be valid for a group to be eligible.
Appendix~\ref{app:capture-settings} records the frozen configurations and
their model-identity checks.

The adapter recorded ordered tool names, arguments, returned synthetic values,
and result hashes. All 16 groups had
\(\dar=\tarstrong=1\). Qwen called all three tools on every case. Granite
called all three on tool-required cases and only the policy tool on
self-contained cases. Neither configuration produced an empty path; live
inference therefore did not exercise the intended no-tool control.

\paragraph{What we changed after execution.}
Each predeclared transformation preserved a record, removed its first recorded
call, or removed all its recorded calls. Decisions and the dispatcher-side
completion record stayed fixed. The dispatcher records completion before the
adapter constructs its trace. This isolates omissions introduced into that
later record. There are \emph{48 executions and 144 paired record views},
not 144 independent model trials.

\begin{table}[htbp]
\centering\small
\caption{Controlled omissions in paired views of the same executions.
Each row contains 48 episode views. Strong TAR is calculated before checking
capture consistency. Flags identify injected record omissions, not harmful
agent actions.}
\label{tab:capture-loss}
\begin{tabularx}{\textwidth}{@{}Xrrrr@{}}
\toprule
Recorded view & Omissions & \(\tarstrong\) & Name flags & Dispatch flags \\
\midrule
Unchanged & 0/48 & 1.0 & 0/48 & 0/48 \\
Drop first call & 48/48 & 1.0 & 24/48 & 48/48 \\
Drop all calls & 48/48 & 1.0 & 24/48 & 48/48 \\
\bottomrule
\end{tabularx}
\end{table}

Every transformed group still has perfect decision and strong-path agreement
(Table~\ref{tab:capture-loss}). A schema check also accepts every view:
the lists remain well formed, including when empty. Checking for required
tool names flags the 24 tool-required episodes in each corrupted view. It
misses the other 24 because their optional calls are absent from the required
name list. Exact comparison with the dispatcher record flags all injected
omissions and flags no unchanged record. Requiring that consistency before
scoring makes the corrupted groups unscorable. An ordinary trace comparison
with the same observations achieves the same result.

\paragraph{What the diagnostic establishes.}
The two streams share a process and tool implementation. Their comparison
detects the injected record omissions; a shared omission or unrecorded
external effect can escape both. The detection counts give no estimate of
deployment prevalence. Hashing a returned value establishes neither its source
nor its semantic correctness. Independent observation of execution remains
untested.

A separate, post hoc check found one stable violation of the synthetic task
rule. Granite returned \texttt{INVESTIGATE} on all three \texttt{CAP-08}
repeats for a 2,000-cent difference, although the rule requires
\texttt{ESCALATE} at 1,000 cents. Both agreement scores remained perfect.
The repeated decision violates the explicit synthetic rule, giving a concrete example of agreement without rule conformance.

%% file: banking.tex
\section{Evidence before execution}
\label{sec:banking}

The replay experiments ask whether an agent repeats its decisions and recorded
work. A second question concerns permission: \emph{does the evidence available
to the agent justify its next action?} A banking assistant can know that a
referral offer expires on a given date without knowing today's date. Repeating
\allow{} does not supply the missing fact. Blocking every referral avoids that
particular unsupported action, but also prevents permitted work. We therefore
measure permission, task completion and cost separately.

\subsection{What the banking experiment changes}
We use the external $\tau$-Knowledge banking environment
\citep{shi2026tauknowledge}. Its pinned source contains 97 tasks and 698
knowledge documents; the primary condition supplies task-relevant documents
through \texttt{golden\_retrieval}. All three arms share the task environment,
user simulator, evaluators and structural checks for callable arguments,
mutation metadata and policy presence:
\begin{itemize}
\item \textbf{A: structural checks.} Execute proposals that pass these checks.
\item \textbf{B: generative gate.} Also ask the generator model for a
schema-constrained JSON decision before executing an eligible mutation.
\item \textbf{C: typed-choice gate.} Substitute the Jev API, which returns a
category and class probabilities, for B's judgment.
\end{itemize}
Primary generation and B gating use DeepSeek v4.1 Flash through DeepInfra.
Gemini 3.8 Flash through Google AI Studio supplies the user simulator and
the benchmark's language-based graders. C uses \texttt{typesafe/jev-1.13-20260917}.
An exploratory condition also uses Gemini for generation and B gating;
another replaces supplied retrieval with BM25. These are comparisons of
implemented bundles of models, input limits, parsing and recovery. They do
not isolate a causal effect of model architecture.

Both gates receive delivered policy, visible dialogue and tool results, the
resolved proposal and the generator's textual rationale fields. Hidden grading
criteria, reference actions and observer snapshots are excluded. The ordered
rubric is: \block{} if evidence establishes a disqualifier; otherwise
\review{} if required evidence is missing; otherwise \allow{} if the
prerequisites are supported. For example, knowing an offer has expired supports
\block{}, whereas lacking the current date supports \review{}. Proof supplied
by a customer must retain its own provenance even if a database fact agrees.
The two gates share an information schema, but earlier interventions can make
their later dialogues and proposals different.

Only eligible assistant mutations cross this boundary. Reads, spoken
disclosures and user actions remain outside it. Both \block{} and
\review{} withhold execution, return the same generic feedback and permit at
most two further proposals; a third denial exhausts recovery. Here,
``recovery'' means another opportunity to propose an action. It does not mean
human review or restoration of durable financial state.

Two input limits matter. A common 90,000-byte gate-input limit converts an
oversize state into operational \review{} before either judge is called.
It is a byte heuristic, not a guarantee of equal token capacity. A separate
500,000-byte generator-request limit can reject an episode before a request
is sent. The recorder distinguishes these events from categorical decisions,
then records dispatch, tool results and effects. Database and action graders
see executed calls rather than denied proposals; communication graders retain
the original dialogue. We call the external benchmark's task-success outcome
\emph{native success}. It is not an independently adjudicated measure of
action-level policy compliance.

\subsection{Completion: the full schedule and the usable outcomes}
Six development tasks were excluded from the environment's 97 tasks. Seeded
sampling selected 90 of the remaining 91 for the primary schedule, with three
trials per arm and 810 episodes; the unused task was not a replacement.
The exploratory Gemini and BM25 schedules use 30-task and 60-task subsets
of those 90, adding 90 and 180 episodes, respectively. Their subsets overlap
on 15 tasks. Appendix~\ref{app:banking-selection} records IDs and seeds.
Each task receives a randomized arm order over fixed trial
bundles, and a separate randomization determines execution order. The three
cohorts remain separate throughout analysis. All 1,080 scheduled episodes
produce terminal records, but only 1,033 have evaluable native outcomes;
47 remain unknown (Table~\ref{tab:banking-native}). A completed record is
therefore not necessarily a usable success or failure observation.

\begin{table}[!htbp]
\centering
\small
\caption{Native outcomes, preserving every scheduled episode. Each arm entry
is \emph{successes / known outcomes / unknown outcomes}. Tasks $\times$ trials
is the scheduled count per arm. Primary and exploratory cohorts are not pooled.}
\label{tab:banking-native}
\begin{tabular}{@{}llrrrr@{}}
\toprule
Role & Generator / retrieval & Tasks $\times$ trials & A & B & C\\
\midrule
Primary & DeepSeek / golden & $90\times3$ & 121/260/10 & 81/263/7 & 70/263/7\\
Exploratory & Gemini / golden & $30\times1$ & 15/29/1 & 13/29/1 & 8/30/0\\
Exploratory & DeepSeek / BM25 & $60\times1$ & 15/47/13 & 8/57/3 & 9/55/5\\
\bottomrule
\end{tabular}
\end{table}

The outcome rule distinguishes a scored failure from an unavailable result.
A generator or simulator reaching its output limit with no usable text or
tool call is scored as failure on its valid prefix. Other malformed responses
and communication failures leave the outcome unknown. The planned primary
comparison required complete B/C results across the selected tasks. With
that requirement unmet, we report observed differences, uncertainty estimates,
and bounds on what the missing outcomes could change.
Appendix~\ref{app:banking-missingness} gives the exact testing requirements
and analysis populations.

Among the 77 primary tasks with all three trials known in both B and C, C's
mean success rate is 3.03 percentage points lower, with a descriptive 95\%
task-bootstrap interval of $[-7.79,+1.73]\pp$. Each task has equal weight;
its repeats stay together. Comparisons with A also favor A on their complete
task subsets (Appendix~\ref{app:banking}).

We can additionally ask what the missing outcomes could change on the
\emph{collected schedule}. Even if all seven unknown C episodes succeeded,
C would reach only 77 successes, below B's 81 already known successes. B
similarly cannot reach A. Thus the ordering A$>$B$>$C survives every binary
assignment of the primary unknown outcomes. C minus B lies between
$-6.67$ and $-1.48\pp$. These logical bounds describe missing-outcome
ambiguity on this schedule; the bootstrap interval describes a selected
complete-task subset. They are different statements, so the latter's
inclusion of zero is not a contradiction.

The exploratory Gemini schedule also retains A$>$B$>$C under every
missing-outcome assignment. BM25 retains A above both semantic bundles,
but its B/C order is unresolved. Eighteen of BM25's 21 unknown outcomes
come from the 500,000-byte generator-request limit. Complete-case
comparisons therefore select materially different populations. Neither the
ordering nor the limits establish which arm produces safer actions: lower
native completion could reflect useful intervention, mistaken denial or
failure to obtain a decision.

\begin{figure}[t]
\centering
\includegraphics[width=\linewidth]{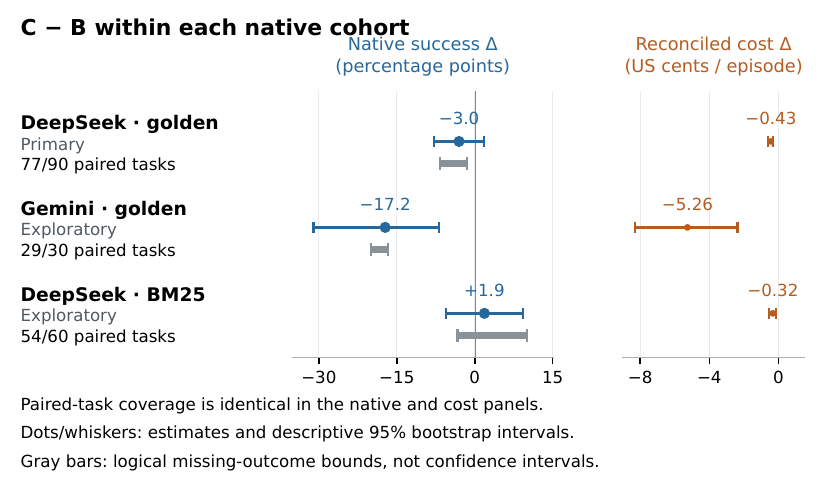}
\caption{Typed-choice C minus generative B within each banking cohort.
Points and whiskers show complete-task means and descriptive 95\%
task-bootstrap intervals; both panels use the same task pairs. Gray bars
bound completion differences across the full scheduled cohort by assigning
each unknown outcome both possible values. Negative cost means less
expenditure; negative success means less completed work. The primary
completion analysis uses 5,000 bootstrap draws and costs use 2,000.}
\label{fig:banking-cost}
\end{figure}

\subsection{A valid answer is not the same as a covered action}
The primary traces contain 916 eligible mutation contexts for B and 864 for
C, including prefixes whose final task outcomes are unknown. These are
dependent observations from different trajectories, not matched action pairs.
The input-size check excludes 259 B contexts and 222 C contexts before inference.
Among returned answers, categorical validity is 644/656 (98.2\%) for B and
638/641 (99.5\%) for C. Against \emph{all eligible contexts}, however,
valid-decision coverage is only 644/916 (70.3\%) and 638/864 (73.8\%).
Each arm also has one context without a response or verdict. Reporting
returned-answer validity alone omits most of the unavailable decisions.

A trace audit links all 71 B and 108 C recovery-exhaustion terminations
to a third denied proposal. In 67 B and 57 C episodes, every denial came
from input admission without a judge call. Only one B and 41 C episodes
exhausted recovery entirely through semantic categorical decisions. The
audit identifies how these episodes stopped; it cannot tell us which would
have succeeded with a larger envelope or a different recovery policy.

\subsection{Lower expenditure can accompany less completed work}
Whole-episode costs cover every request role and failed attempt. Reconciled
banking expenditure is \$27.2458; 1,047/1,080 episodes have usable cost traces
and 33 do not. Matched comparisons exclude eight review-inferred zero-dollar
settlements. Appendix~\ref{app:banking} explains these accounting limits.

C costs less than B on jointly complete task pairs: mean episode differences
are $-\$0.004345$ for primary DeepSeek (77/90 pairs), $-\$0.052625$ for
Gemini (29/30), and $-\$0.003245$ for BM25 (54/60). The primary descriptive
95\% cost interval is $[-\$0.005782,-\$0.003068]$. Figure~\ref{fig:banking-cost}
places these costs beside completion. Early stopping or fewer recoveries can
reduce spending while leaving less work done. A cost-per-success ratio would
have a different denominator; a cost-per-policy-compliant-success measure
would additionally require independent action labels that this study lacks.
These are observed API charges, excluding human review and production
operating costs.

\section{Gate diagnostics: what repeated decisions miss}
\label{sec:probes}

Native interventions change later dialogue and actions. To inspect narrower
questions, we also query B and C on fixed visible states using the primary
judge configuration. These diagnostics execute no new banking actions.
Their 648 constructed-rule, 1,056 natural-context and 240 selected-case
requests are related states and repeats, not 1,944 independent tasks or
replications of the native experiment.

\subsection{Constructed policy states: consistently wrong is still wrong}
Six families cover identity evidence, verification timing, a documented
bypass, exact email-change intent, referral expiry and referral terms. Six
scenarios per family each have permitted, forbidden and missing-evidence
variants: 36 related triplets, or 108 states. Each is queried three times
per gate. Labels are constructed from explicit rules outside the model
input. Appendix~\ref{app:banking-rules} specifies the rule predicates and
gives a retained example triplet.

For a state with three available decisions, modal agreement is the fraction
in its most frequent category. We average this over complete state groups.
Exact label match instead asks whether each available decision matches the
constructed policy label. Table~\ref{tab:banking-rules} shows why the
distinction matters: C repeats its verdict more often, yet matches the
labels less often, including errors repeated on all three requests.
Assigning every unavailable decision either
match outcome bounds full-schedule match at 80.6--83.0\% for B and
54.0--54.9\% for C. Missing decisions cannot reverse that finite-bank
separation.

\begin{table}[!htbp]
\centering
\small
\caption{Constructed policy-rule results. Decision rates use qualified
requests; modal agreement uses complete three-repeat state groups. Only
98 states are complete for both gates. Provider/cache freshness is
unqualified, so agreement describes observed repetition.}
\label{tab:banking-rules}
\begin{tabular}{@{}lrr@{}}
\toprule
Measure & B: generative & C: typed choice\\
\midrule
Qualified / scheduled & 316/324 & 321/324\\
Exact rule-label matches & 261/316 (82.6\%) & 175/321 (54.5\%)\\
Modal agreement; complete groups & 93.1\%; 101/108 & 98.7\%; 105/108\\
Unanimously wrong groups & 9 & 46\\
\midrule
\allow{} with missing evidence & 6/107 & 36/108\\
\review{} with missing evidence & 56/107 & 2/108\\
\allow{} when forbidden & 2/103 & 0/106\\
\allow{} when permitted & 106/106 & 67/107\\
\block{} when permitted & 0/106 & 40/107\\
\bottomrule
\end{tabular}
\end{table}

C's zero forbidden-approval count covers only one part of the result: it
also approves missing-evidence proposals and blocks permitted proposals.
Exact-label errors include
\block{}/\review{} substitutions, which both suppress execution here.
Label error is therefore neither a harm rate nor an estimate of policy
accuracy on natural traffic.

\subsection{Natural contexts: sensitivity to how the same input is written}
The natural-context probe varies tool-call identifiers and JSON presentation
across 44 gate contexts. Four representations, three repeats and two gates
produce 1,056 requests. B returns 523/528 usable decisions and C 525/528.
Selection rules, transformations and the statistic are specified in
Appendix~\ref{app:banking-invariance}.

We measure decision changes under the transformations above each gate's own
raw-repeat baseline, then compare C minus B at task level. Identifier
renaming gives $+0.88\pp$ on 38/44 complete tasks; syntax changes give
$+8.33\pp$ on 40/44; both give $+10.00\pp$ on 40/44. Eight unavailable
decisions make the separate prespecified test family unavailable. With no
policy-accuracy labels, this diagnostic measures sensitivity to representation rather than
which judge is correct.

\subsection{One referral case: missing dates and retained rationale}
In a selected development task, both B and C allow a referral-capability
grant, the simulated user ends satisfied, and native database reward is zero.
Their trajectories differ before gating and neither gate denies a proposal.
That observation cannot identify a gate effect. Instead, we take two
related visible gate contexts from the task and vary the evidence they contain.

Each context has three timing states (date absent, before expiry, after
expiry) and two rationale conditions (retained, removed). Two judges and
ten repeats over these 12 visible states produce 240 requests, of which
218 are usable. Only the dated before/after pair changes the date alone.
Adding a date to the missing-date context also adds a tool-call scaffold;
removing the generator's two \texttt{reasoning\_details} fields changes
both information and length.

C allows all 40 requests lacking a date. After expiry, C allows 18/20
requests with the generator rationale and 1/20 without it. B blocks all
40 after-expiry requests. B's missing-date cells contain 20 allows, three
reviews and 17 unavailable decisions; another five unavailable decisions
occur before expiry. All 22 are B output-limit failures. Omitting them
would selectively remove difficult cases from B's results.

\begin{figure}[!htbp]
\centering
\includegraphics[width=0.94\linewidth]{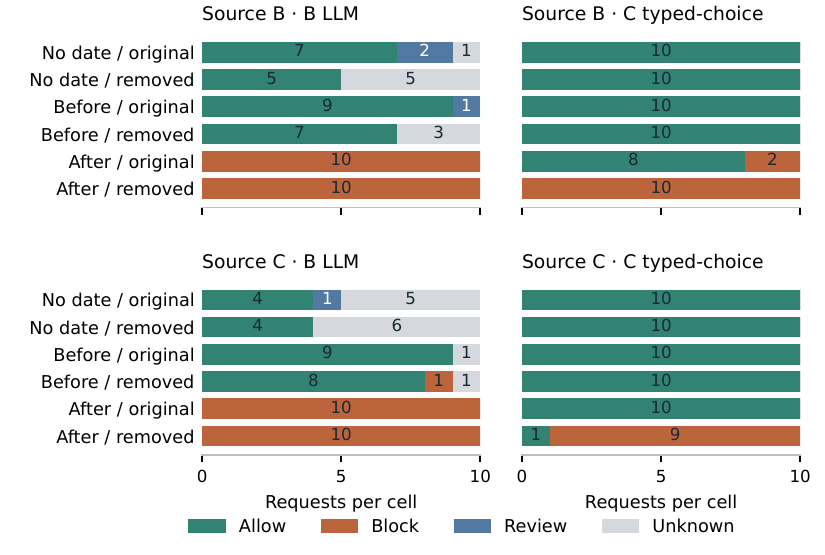}
\caption{Selected referral diagnostic, retaining 240 requests over 12
visible states. Each of the 24 judge--state cells contains ten requests,
including unavailable decisions. The two source contexts come from one
development task. Retained rationale accompanies more after-expiry C
approvals (18/20 versus 1/20); its removal changes information and length.
No revised gate is tested in native business execution.}
\label{fig:banking-referral}
\end{figure}

Figure~\ref{fig:banking-referral} preserves all requests. Missing-date
\review{} follows the rubric; post-expiry \block{} follows the temporal
rule. Other prerequisites remain unresolved before expiry, so that
condition has no unconditional \allow{} label. The result establishes
sensitivity to retained rationale on this selected task. It does not identify
persuasion, self-justification or an internal reasoning mechanism. Textual
rationale is observable output, not verified access to computation
\citep{jacovi2020faithfulness}.

%% file: discussion.tex
\section{What agreement leaves unmeasured}
\label{sec:discussion}

The studies separate three kinds of evidence that are easy to conflate. Replay agreement describes the behavior visible in a record. The capture diagnostic checks whether that record preserves the observed execution. The banking probes compare a decision with the evidence and rule that should govern it. Each test answers a question left open by the preceding measurement; none requires treating the separate studies as one pooled experiment.

The prospective replays show why the level of recording matters. Adding canonical arguments reveals differences behind matching tool names on the same episodes. The cancellation example identifies a second loss of information: averaging signed gaps can hide a case that is visible in the individual results. Together, these findings favor retaining the case-level comparison and the changed fields alongside the aggregate score.

The deletion diagnostic goes one step earlier. All repeats can contain the same incomplete trace, so agreement within the traces cannot expose the omission. Comparison with tool-completion records detects the injected loss. This is an observation requirement demonstrated by a controlled counterexample; ordinary record comparison supplies the detection mechanism.

The banking study shows why those distinctions matter when a judgment controls execution. High validity among returned verdicts coexists with much lower coverage of all eligible proposals, because many inputs never reach a judge. A verdict can also repeat while disagreeing with an explicit rule. Finally, the lower-cost typed-choice bundle completes fewer primary tasks than the generative bundle under every assignment of missing outcomes. The useful unit of evaluation is therefore the full implemented intervention, including input-size checks and recovery. The stopping records identify changes worth testing, but their effects on completion and action correctness would require a further intervention.

\input{audit_protocol}

\section{Relation to existing evaluation}
Financial benchmarks such as FinBen, FinAgentBench, and FinResearchBench assess task capability or financial-research quality \citep{xie2024finben,choi2025finagentbench,sun2025finresearchbench}. FinTrace evaluates financial trajectories against expert annotations, while TRAJECT-Bench assesses tool use against reference trajectories \citep{cao2026fintrace,he2026traject}. These reference-based targets are needed to interpret whether a changed path is acceptable. DFAH's replay analysis instead compares executions of the same case and measures the variation that a final decision hides.

$\tau$-bench studies repeated task success, ToolSandbox evaluates stateful milestones, and $\tau$-Knowledge supplies the banking tasks used here \citep{yao2024taubench,lu2025toolsandbox,shi2026tauknowledge}. Our intervention adds a separate accounting of proposed actions, unavailable gate judgments, denials, executed calls, and task outcomes. That accounting explains why a high rate of valid model responses can coexist with limited action coverage and low completion.

Studies of consistency and reasoning stability examine repeated outputs \citep{wang2025consistency,potamitis2025reasonbench,wang2023selfconsistency}; work on agent visibility and evaluation schemas concerns the observations retained for inspection \citep{chan2024visibility,batzner2026every}. The cancellation and deletion examples connect these concerns by showing what aggregation and incomplete recording can hide. They are diagnostic counterexamples rather than new statistical or detection algorithms. Runtime-governance and enforcement work addresses the conditions under which actions may proceed \citep{szpruch2026runtime,schneider2000enforceable,shlomov2026governance}. Our banking comparisons characterize the behavior and costs of specific implemented gates; the rule probes supply narrower reference checks on their verdicts.

\section{Limits and next evidence needed}
\label{sec:limitations}

\paragraph{Correctness and financial significance.}
The replay fixtures and policy states are authored, and the banking study uses one synthetic environment. Independent expert assessment of action correctness and financial significance remains outstanding. The study-specific differences and constructed-rule errors therefore support diagnosis, while their prevalence and consequences in financial operations remain open. Natural-context representation probes have no policy-accuracy labels, and the proposed review categories have not been validated as risk thresholds.

\paragraph{Which comparisons the designs support.}
Historical configurations differ in coverage, providers, sampling, repeat counts, and harness behavior; the separate prospective studies also differ in workflows and capture depth. Banking arms combine models, APIs, input limits, parsing, and recovery, with later contexts changing after intervention. The selected rationale deletion changes both information and input length. These designs support configuration and system comparisons, not attribution to model architecture or hidden reasoning.

\paragraph{Observation and missing data.}
Missing records or outcomes can change which cases enter a comparison, especially when larger requests are more likely to be unavailable. The reported intervals describe the retained cases; the banking bounds ask what the unknown outcomes could change on the scheduled tasks. Neither establishes performance on future customer traffic. Appendices~\ref{app:prospective-coverage} and~\ref{app:banking-missingness} specify the coverage and testing rules. Historical parser substitutions lack event-level provenance, the capture reference shares a process with the recorder, and provider cache freshness and per-request model weights remain unverified.

\paragraph{Operational scope.}
The gate acts only on eligible assistant mutations. Reads, disclosures, and user actions lie outside that boundary. Denials lead to bounded automated retries, with no completed human-resolution workflow. The measured charges cover API use, excluding human review and production overhead. Independent action labels, a separately observed execution reference, and matched interventions on admission and recovery are the next tests needed to assess operational value.

\paragraph{Research process.}
Protocols were specified in internal study plans rather than publicly preregistered. Interim banking outcomes were visible during collection, so the analyses were not blinded. Model assistance was used in method review and manuscript development.

\section{Conclusion}
DFAH-Bench examines how much evidence agreement provides about a financial agent. The replays expose changed tool paths behind repeated decisions; the capture diagnostic shows how systematic omissions survive an agreement check; the banking studies separate repeated verdicts from rule match, action coverage, and completed work. The resulting evaluation keeps the changed record, the rule governing an action, and the downstream outcome visible together. The practical contribution is a way to identify what an agreement score leaves unresolved and choose the next test accordingly. A change in tool use calls for a different investigation from a missing record, an unauthorized action, or unfinished work. Keeping those distinctions visible ties claims of reliability to specific observations and helps researchers identify what an agent or its surrounding system needs to improve.

%% file: audit_protocol.tex
\subsection{An audit procedure motivated by the findings}
\label{sec:audit}

The experiments suggest the following procedure. Its operational value and review costs have not been tested.
\begin{enumerate}[leftmargin=*,itemsep=3pt]
\item \textbf{Define what is held fixed.} Record case and fixture versions, model and provider identities, sampling settings, tool implementations, required observations, and any permitted no-tool paths.
\item \textbf{Account for every planned execution.} Preserve terminal status, parsing failures, fallback labels, and capture eligibility. Compare adapter traces with tool-completion observations before calculating agreement; an unresolved record stays unresolved.
\item \textbf{Inspect the changed fields.} Report case-level decision and path agreement alongside the averages. Count variable paths among unanimous-decision cases, and separately report the share of all cases flagged for review.
\item \textbf{Apply the task's rule.} Review operations, arguments, evidence, and state. A proposed rubric distinguishes documented benign variation, possible control failure, and an unresolved case. These dispositions require independent task-specific judgment rather than automatic assignment from a path difference.
\item \textbf{Follow the intervention through completion.} Keep permission, dispatch, task outcome, and cost separate. Include judgments unavailable before inference, exhausted retries, and the time spent reviewing exceptions.
\end{enumerate}

%% file: replay_appendix.tex
\section{Historical corpus construction}
\label{app:revision}

The original replay corpus combined three fixtures.
Table~\ref{tab:lineage} traces the exclusions from raw records to the
retained primary sample. The reasons and sensitivity checks follow below.

\begin{table}[htbp]
\caption{Construction of the historical replay sample.}
\label{tab:lineage}
\centering
\small
\begin{tabular}{@{}lrr@{}}
\toprule
Stage & Episodes & Groups \\
\midrule
Raw replay records & 8,129 & -- \\
After singleton removal & 8,127 & 1,338 \\
Remove portfolio fixture (2,612 episodes) & 5,515 & 889 \\
Remove two-case DeepSeek pilot (14 episodes) & 5,501 & 887 \\
Retain configurations with observed tool use & 4,157 & 719 \\
\bottomrule
\end{tabular}
\end{table}

The required-channel audit separately inspected 5,502 non-portfolio,
non-DeepSeek records before replay-group qualification and found no missing
decision or trajectory fields.  One DataOps record belonged to a singleton
group and was therefore ineligible for replay analysis, leaving the 5,501
episodes above.

\paragraph{Sensitivity to the no-tool exclusion.}
The post-run configuration-level rule excludes 1,344 episodes in 168 groups:
800 episodes in 100 Granite 3.3 groups and 544 episodes in 68 Mistral 7B
groups. Their trajectory fields are present but contain no observed tool
calls. Including these groups leaves the 122 unanimous-decision groups with
varying paths unchanged, while increasing the unanimous-decision denominator
from 627 to 791. The pooled descriptive rate therefore changes from
$122/627=19.5\%$ to $122/791=15.4\%$. Configuration-specific results are
unchanged. This sensitivity applies to the observed configuration--harness
combinations.

Portfolio-derived results were removed because authored totals, distributions,
rule tags and labels were inconsistent. Historical evidence contacts were also
removed: re-logging disproportionately overwrote outputs from decision-divergent
groups, and surviving hashes identify serialized outputs rather than contacted
sources. Filtering this irreversible, decision-dependent loss would change the estimand.

The historical identifier \path{claude-opus-4-20250514} denotes Claude Opus 4;
Opus 4.5 has a different identifier. Both Claude configurations omitted
temperature in their requests despite zero-valued metadata, warranting their
provider-default sensitivity designation.

\section{Configuration and Canonicalization Contract}
\label{app:contract}

The replay unit is keyed by task, case, configuration, and replay index.
Comparability requires equality of the suite version, fixture hash, model and
provider identifiers, tool schemas, system and user-message hashes, decoding
controls, turn and token limits, adapter version, and every channel marked
required.  A different suite version is non-comparable unless an analysis
explicitly opts into a cross-version comparison.

In the prospective API and local reconciliation extensions, JSON values are
canonicalized recursively: object keys are sorted;
arrays retain order; JSON types are preserved; separators are compact; text is
UTF-8; and non-finite or non-JSON values are rejected.  The hashed byte string
ends with a newline. The separate September 6 capture diagnostic hashes its
compact, key-sorted JSON without a trailing newline, following its retained
collector. These cohort-specific encodings are preserved; their result hashes
are not compared across cohorts. We define three path projections:

\begin{align}
P^{\mathrm{seq}}_i
  &= (n_{i1},\ldots,n_{im_i}),\\
P^{\mathrm{arg}}_i
  &= ((n_{i1},a_{i1}),\ldots,(n_{im_i},a_{im_i})),\\
P^{\mathrm{res}}_i
  &= (h(r_{i1}),\ldots,h(r_{im_i})),
\end{align}
where \(n\) is the tool name, \(a\) its canonical arguments, and \(h(r)\) the
hash of the canonical deterministic result.  The strong path is the ordered
triple \((n,a,h(r))\).  Including the tool name in the argument projection
prevents equal argument objects passed to different tools from collapsing.

For every projection \(x\), agreement is the modal share over the same
eligible replay set:
\[
  A_x(c)=\frac{\max_v\sum_{i=1}^{N_c}\mathbf{1}[x_i=v]}{N_c}.
\]
Thus \(\dar=A_D\), \(\tarseq=A_{P^{\mathrm{seq}}}\), and
\(\tarstrong=A_{P^{\mathrm{strong}}}\).  Gaps are paired within case before
task aggregation.  They may be negative because the decision and path modes
are computed separately; a negative value means the selected path projection
is more concentrated than the decisions.

\section{Historical Configuration and Task Results}
\label{app:task-results}

\begin{table}[!htbp]
\caption{Retrospective replay agreement.  Panel A contains the six
configurations with complete 100-group coverage; Panel B retains the two
contiguous-prefix configurations separately.  Replays is the number of
identical executions per available case.  Tool is the percentage of episodes
making at least one tool call.  Brackets are paired case-resampling 95\%
sensitivity intervals for \(\gapmetric\), in percentage points.  $\dagger$
marks the provider-default sensitivity configurations.}
\label{tab:results}
\centering
\small
\setlength{\tabcolsep}{3.5pt}
\resizebox{\linewidth}{!}{%
\begin{tabular}{lrrrrrrrl}
\toprule
Configuration & Tasks & Groups & Episodes & Replays & Tool & DAR & \(\tarseq\) & \(\gapmetric\) [95\%] \\
\midrule
\multicolumn{9}{l}{\textit{Panel A: complete 100-group coverage}} \\
Qwen 3.5 & 2 & 100 & 800 & 8 & 100.0 & 100.0 & 100.0 & 0.0 [0.0, 0.0] \\
Gemma 4 & 2 & 100 & 800 & 8 & 99.0 & 99.9 & 99.8 & 0.1 [0.0, 0.4] \\
Qwen 2.5 7B & 2 & 100 & 800 & 8 & 100.0 & 99.6 & 99.6 & 0.0 [$-$0.4, 0.4] \\
GPT-OSS 20B & 2 & 100 & 800 & 8 & 98.9 & 98.0 & 97.9 & 0.1 [$-$0.8, 0.9] \\
Gemini 2.0 Flash & 2 & 100 & 300 & 3 & 99.3 & 93.7 & 88.7 & 5.0 [1.7, 8.7] \\
Claude Sonnet 4$\dagger$ & 2 & 100 & 300 & 3 & 100.0 & 94.3 & 73.3 & 21.0 [16.3, 25.7] \\
\addlinespace[3pt]
\multicolumn{9}{l}{\textit{Panel B: limited contiguous-prefix coverage}} \\
Gemini 2.5 Pro & 1 & 44 & 132 & 3 & 95.5 & 88.6 & 75.0 & 13.6 [6.8, 20.5] \\
Claude Opus 4$\dagger$ & 2 & 75 & 225 & 3 & 100.0 & 89.0 & 70.3 & 18.7 [12.3, 24.7] \\
\bottomrule
\end{tabular}}
\end{table}

Table~\ref{tab:task-results} exposes the task composition hidden by the compact
configuration table.  It retains the historical tool-name projection; the
later strong API protocol with arguments and results is reported separately.

\begin{table}[htbp]
\caption{Retrospective task-level agreement (percent).  Pro has a 44-case
compliance prefix; Opus has complete compliance and a 25-case DataOps prefix.
All other cells contain 50 cases. Gaps are rounded to the nearest tenth,
with ties to even.}
\label{tab:task-results}
\centering
\small
\renewcommand{\arraystretch}{1.12}
\begin{tabular*}{0.94\linewidth}{@{\extracolsep{\fill}}llrrr@{}}
\toprule
Configuration & Task & DAR & \(\tarseq\) & Gap \\
\midrule
Qwen 3.5 & Compliance & 100.0 & 100.0 & 0.0 \\
Qwen 3.5 & DataOps & 100.0 & 100.0 & 0.0 \\
Gemma 4 & Compliance & 99.8 & 99.5 & 0.2 \\
Gemma 4 & DataOps & 100.0 & 100.0 & 0.0 \\
Qwen 2.5 7B & Compliance & 100.0 & 100.0 & 0.0 \\
Qwen 2.5 7B & DataOps & 99.3 & 99.3 & 0.0 \\
GPT-OSS 20B & Compliance & 98.0 & 99.0 & -1.0 \\
GPT-OSS 20B & DataOps & 98.0 & 96.8 & 1.2 \\
Gemini 2.0 Flash & Compliance & 89.3 & 88.7 & 0.7 \\
Gemini 2.0 Flash & DataOps & 98.0 & 88.7 & 9.3 \\
Claude Sonnet 4 & Compliance & 94.0 & 79.3 & 14.7 \\
Claude Sonnet 4 & DataOps & 94.7 & 67.3 & 27.3 \\
Gemini 2.5 Pro & Compliance & 88.6 & 75.0 & 13.6 \\
Claude Opus 4 & Compliance & 90.0 & 78.0 & 12.0 \\
Claude Opus 4 & DataOps & 88.0 & 62.7 & 25.3 \\
\bottomrule
\end{tabular*}
\end{table}

\begin{figure}[htbp]
\centering
\includegraphics[width=0.64\columnwidth]{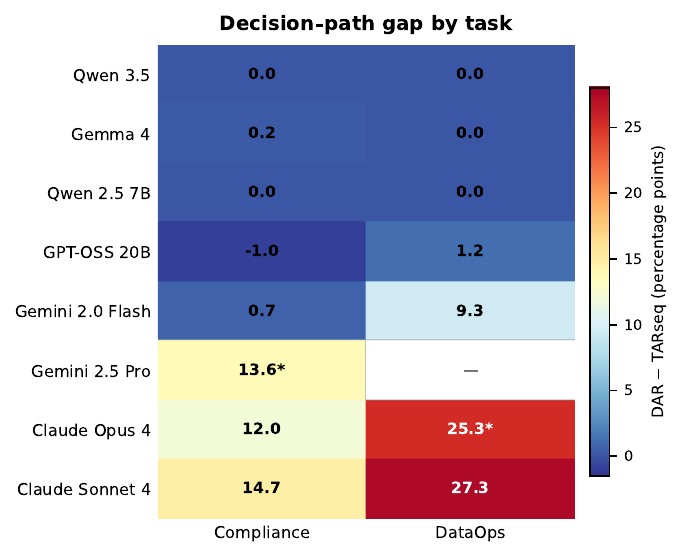}
\caption{Corrected task-level decision--path gaps for the retained fixtures.
Values are percentage points; an asterisk marks the
44-case Gemini Pro compliance prefix or the 25-case Opus DataOps prefix.
Blank cells were not observed.}
\label{fig:task-gap}
\end{figure}

Removing one case at a time leaves Flash's task-weighted gap between 4.4 and
5.3 points.  The larger
DataOps gap is therefore distributed across the authored set, though the
exercise remains conditional on these cases and three observed replays; the
task-level decomposition is shown in \autoref{fig:task-gap}.

\FloatBarrier
\subsection{Exploratory Decision Concentration}
\label{app:concentration}

\begin{figure}[htbp]
\centering
\includegraphics[width=0.96\columnwidth]{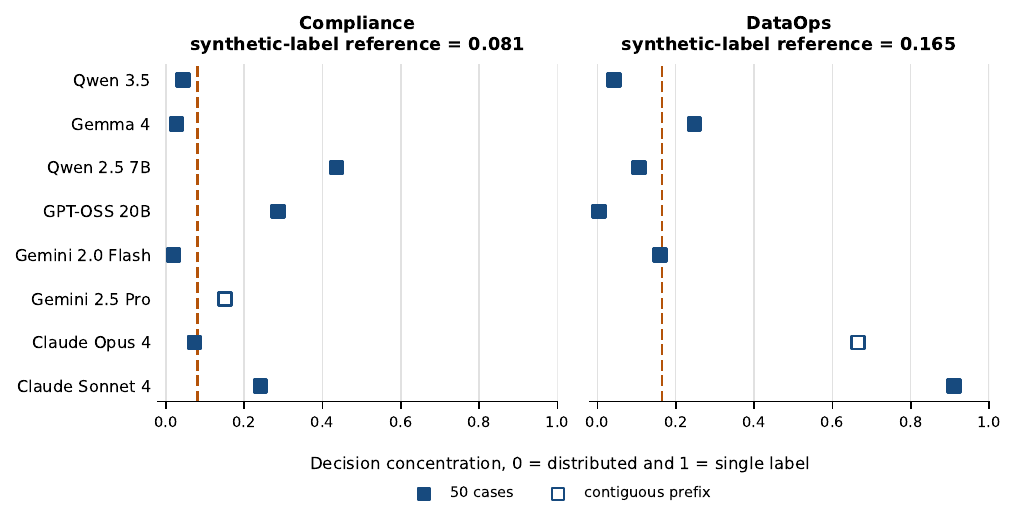}
\caption{Modal-decision concentration across cases in the corrected
compliance and DataOps slice.  Dashed lines are the concentrations of the
provisional synthetic fixture labels.  Filled
markers cover all 50 cases; open markers denote contiguous prefixes.}
\label{fig:concentration}
\end{figure}

The DAR--TAR gap is a within-case replay measure.  A separate question is
whether a configuration gives nearly the same label to many different cases.
For a task with three labels and modal-label distribution \(\mathbf{p}\), we
report
\[
  \mathrm{DCB}=1-\frac{H(\mathbf{p})}{\log 3}.
\]
DCB is zero for an even distribution and one when every case receives one
label.  It describes the distribution of decisions across cases. Version 1 used DCB to
name three behavioral profiles; v2 drops that taxonomy and shows the statistic
by task instead; the resulting concentrations are shown in
\autoref{fig:concentration}.

The largest value is 0.911 for Sonnet on DataOps: 49 of 50 modal decisions are
\emph{quarantine}, compared with a synthetic-label concentration of 0.165.
This identifies an output-distribution pattern for substantive review.
Open-prefix points describe only their observed case subsets.

\section{Replay-Count and Parser Sensitivities}
\label{app:sensitivity}

For each of the four eight-replay configurations, we drew 500 fixed-seed
three-replay subsets per case and recomputed the task-weighted paired gap.
Table~\ref{tab:n3} shows that the near-zero local pattern persists at the matched
replay count.  The subsamples reuse the observed runs.

\begin{table}[htbp]
\caption{Three-replay subsampling of eight-replay historical rows.  Values are
median [2.5th, 97.5th percentile] gap in percentage points.}
\label{tab:n3}
\centering
\small
\begin{tabular}{@{}lc@{}}
\toprule
Configuration & Three-replay gap \\
\midrule
Qwen 3.5 & 0.0 [0.0, 0.0] \\
Gemma 4 & 0.0 [0.0, 0.3] \\
Qwen 2.5 7B & 0.0 [-0.3, 0.3] \\
GPT-OSS 20B & 0.3 [-0.7, 1.0] \\
\bottomrule
\end{tabular}
\end{table}

A within-task sign-flip test over observed per-case gaps gives \(p=0.0142\)
for Gemini 2.0 Flash and \(p=0.0010\) for Gemini 2.5 Pro under 10,000
permutations (seed 42).  This tests sign symmetry/exchangeability of the paired
finite-corpus gaps.

The legacy parser did not mark when it substituted the final ontology label.
Under a conservative adversarial reassignment, the Flash gap remains positive
through a 2\% fallback rate, reaches zero at 16/300 episodes (5.3\%), and
reverses above that point.  The Pro prefix remains positive through 10\% and
reaches zero at 18/132 episodes (13.6\%).  These are stress bounds; the actual
fallback rate remains unobserved.  Prospective parsing fails closed and
records provenance.

\section{Prospective coverage and component comparisons}
\label{app:components}

\paragraph{Record eligibility and coverage.}
\label{app:prospective-coverage}
All 600 scheduled API replay episodes reached terminal states. Ten records
lacked a valid required channel, each in a different three-repeat group.
The rule set before collection required all three repeats to be valid, so
the ten groups and their 30 episodes were excluded. This left 570 episodes
in 190 groups. Sonnet DataOps retained 44 of 50 groups, one below the
planned minimum of 45 for each model/task combination. The shortfall concerns
usable records, not unexecuted runs. We retain the exclusion rule and report
the eligible groups as a descriptive extension.

\paragraph{Comparing record components.}
The prospective API capture retained structured arguments and result objects,
so the strong path can be decomposed offline without another provider call.
The denominator remains the eligible groups above.
Table~\ref{tab:components} reports this post-hoc
projection sensitivity separately from the prespecified strong metric.

\begin{table}[htbp]
\caption{Prospective component agreement and paired gaps (percent).  ``Args''
is ordered tool name plus canonical arguments; ``result'' is the ordered
canonical-result hash. Gaps use unrounded values.}
\label{tab:components}
\centering
\small
\begin{tabular}{@{}lrrrrrr@{}}
\toprule
Configuration & DAR & Names & Args & Result & \(\Delta_{\mathrm{args}}\) & \(\Delta_{\mathrm{result}}\) \\
\midrule
GPT-5.6 Terra & 95.1 & 69.4 & 51.5 & 54.3 & 43.6 & 40.8 \\
Claude Sonnet 5 & 94.2 & 66.9 & 45.0 & 56.9 & 49.2 & 37.2 \\
\bottomrule
\end{tabular}
\end{table}

In all eligible groups, the name--argument projection and the prespecified
strong projection induced the same replay equivalence.  Tool results were
deterministic functions of the fixed fixture and supplied arguments, so adding
their hashes introduced no further split once name and arguments matched.
Result-only agreement is higher because it is a weaker projection: different
calls may serialize to the same result identity.  Task-stratified 95\%
composition intervals are 47.5--55.6\% (Terra) and 41.7--48.3\% (Sonnet) for
name--argument agreement, and 50.2--58.7\% and 52.3--61.7\% for result-only
agreement.  The four corresponding paired gaps have sign-flip
\(p=0.00010\). These are exploratory comparisons of the retained cases.

%% file: capture_appendix.tex
\section{Capture diagnostic: configurations and record encoding}
\label{app:capture-settings}

The September 6, 2026 capture diagnostic used local Qwen 3.5 and Granite 4.1
3B configurations with Ollama 0.33.3 and Q4\_K\_M quantization. Ollama reported
parameter counts of 9.7B and 3.4B, respectively. Requests used temperature
zero, seed 42, disabled thinking, a 4,096-token context, at most 384 generated
tokens per turn, and at most five turns. Eight cases and three planned repeats
per configuration produced the 48 executions reported in
Section~\ref{sec:capture}; these are separate from the historical replay
corpus and other prospective extensions.

Checks before and after each turn matched the server's model inventory to
the frozen manifest digests. Requests still named mutable tags, so these
checks do not establish which weights served an individual request.

Each adapter call record retains its tool name, arguments, returned synthetic
value, and SHA-256 digest of that value's canonical JSON. Compact JSON with
sorted object keys removes whitespace and key-order differences while
preserving types and list order. Strong path agreement compares the ordered
call records. The collector retains the values themselves as well as their
hashes; hashing establishes neither source authenticity nor semantic
correctness.

%% file: banking_appendix.tex
\section{Banking intervention: analysis and reproduction details}
\label{app:banking}

Study-plan and analyzer paths below refer to the author-side research
repository. Appendix~\ref{app:availability} states what is publicly available.

\subsection{Environment, requests and execution records}
The environment is Sierra's \texttt{banking\_knowledge} at commit
\texttt{b7ea9074c1cb}, in the $\tau^3$ implementation repository named
\texttt{tau2-bench}.\footnote{Pinned source:
\url{https://github.com/sierra-research/tau2-bench/tree/b7ea9074c1cba482b30687fecdb5c8425fd6f619}.}
OpenRouter's chat-completions route carries generator, simulator, native
language-grader and B-gate requests. C uses the SystemOne route. Chat
requests restrict providers, disable fallbacks, require parameter support
and retain price caps. Returned identities are checked. Requests are
nonstreaming, without automatic retries or redirects, and use a 120-second
socket timeout. Native chat roles request temperature zero and low
reasoning; C has no equivalent sampling controls. Generation has an
8,192-token output limit; B gating, simulation and grading have 4,096.
These are recorded request settings, not proof of all provider-side behavior.

The boundary separates proposed calls, input admission, gate verdicts,
dispatch, results and effects. Database mutations and in-memory capability
grants have separate hashes; audit writes are separate from business
mutations. Occurrence identifiers distinguish identical requests made more
than once. Request and response hashes, latency and available cache metadata
are retained. Cache freshness and context retention are unqualified.
The 90,000-byte gate envelope also differs from Jev's documented 32k-token
question limit \citep{typesafe2026jevdocs}.

Parsing also changes availability. Three primary C responses contain the
requested choice and a complete probability map summing to 0.99. The
frozen parser's $10^{-4}$ sum tolerance converts them into operational
review. All three episodes ultimately have known failure outcomes. This
is a response-format incompatibility at the validation boundary, not
evidence that their selected categories were semantically wrong.

Database and action evaluators use an execution view excluding denied
proposals and synthetic replies; their scoring criteria are unchanged.
Communication graders use original dialogue. The upstream
\texttt{enforce\_communication\_protocol=false} setting routes mixed
messages to the environment while retaining text for communication scoring.
Partly usable responses retain their classification under the outcome rule
in Section~\ref{sec:banking}. Accounting reconciliation does not
retroactively change outcomes.

\subsection{Task selection}
\label{app:banking-selection}
The pinned inventory contains 97 tasks. Development used
\texttt{task\_005}, \texttt{task\_015}, \texttt{task\_031},
\texttt{task\_048}, \texttt{task\_070} and \texttt{task\_098}, leaving
91 eligible tasks. Uniform sampling using Python's
\texttt{random.Random(seed).sample(sorted(ids), n)} selected 90 primary
tasks (seed 20260920), leaving
\texttt{task\_089} unused. Seeds 20260921 and 20260922 selected the
30-task Gemini and 60-task BM25 subsets of the primary set; those subsets
share 15 task IDs. The retained \path{heldout-task-selection.json} records
selection without held-out outcomes or reference behavior, and forbids
outcome-based replacement, expansion or deletion.

The task schedules and testing hierarchy were fixed before held-out
collection; the probe bank was fixed before its requests were dispatched.

\subsection{Analysis populations and missing-outcome bounds}
\label{app:banking-missingness}
\paragraph{Why the planned tests are not reported.}
The primary two-sided sign-flip test compared C with B at $\alpha=.025$
and required all three trials in both arms to have known outcomes for each
of the 90 selected tasks. Fourteen unknown outcomes affected 13 tasks,
leaving 77 complete task pairs. The prespecified test therefore did not
meet its data requirement. The secondary B--A and C--A tests formed a
Holm(2) family at $\alpha=.025$ that would open only after rejection by
the primary test, so neither is reported as a confirmatory result.
This is a restriction of the chosen analysis plan, not a claim that
incomplete data prohibit all statistical analysis. An analysis restricted
to complete tasks describes a selected subset; its estimates and intervals
below are therefore reported descriptively.

\paragraph{Estimates on complete tasks.}
Let $Y_{iar}\in\{0,1,\bot\}$ be failure, success or unknown for task $i$,
arm $a$ and trial $r$, with $R_i$ scheduled trials per arm for task $i$.
For arms $a,b$, let $I_{ab}$ contain tasks whose trials are all known in
both arms. The complete-task contrast is
\begin{equation}
\widehat\Delta_{ab}=
\frac{1}{|I_{ab}|}\sum_{i\in I_{ab}}
\left(\frac{1}{R_i}\sum_{r=1}^{R_i}Y_{iar}
-\frac{1}{R_i}\sum_{r=1}^{R_i}Y_{ibr}\right).
\end{equation}
Tasks have equal weight and repeats remain in their task cluster. The
primary native bootstrap resamples task differences 5,000 times; matched
costs use 2,000 draws. All reported intervals are descriptive 95\%
intervals.

\paragraph{Bounds across the scheduled tasks.}
With $s_a$ known successes and $u_a$ unknowns among $N$ scheduled episodes
per arm, the realized binary rate lies in $[s_a/N,(s_a+u_a)/N]$, giving
\begin{equation}
\Delta^{\mathrm{schedule}}_{ab}\in
\left[\frac{s_a-s_b-u_b}{N},
\frac{s_a+u_a-s_b}{N}\right].
\end{equation}
Table~\ref{tab:banking-bounds} reports both populations. Bounds enumerate
missing binary assignments; they are not confidence intervals or guarantees
about future tasks. Exploratory native summaries do not serialize their
bootstrap seeds and draw counts, limiting exact reconstruction of displayed
intervals from summaries alone.

\begin{table}[!htbp]
\centering
\small
\caption{Complete-task descriptions and full-schedule logical bounds,
in percentage points. A dash denotes an interval not reproduced here.
Each row retains its own analysis population.}
\label{tab:banking-bounds}
\begin{tabular}{@{}llrrr@{}}
\toprule
Cohort / contrast & Complete tasks & Mean & Descriptive 95\% interval & Schedule bounds\\
\midrule
Primary: C--B & 77/90 & $-3.03$ & $[-7.79,+1.73]$ & $[-6.67,-1.48]$\\
Primary: B--A & 75/90 & $-15.11$ & $[-22.67,-8.00]$ & $[-18.52,-12.22]$\\
Primary: C--A & 78/90 & $-20.09$ & $[-27.78,-13.25]$ & $[-22.59,-16.30]$\\
Gemini: C--B & 29/30 & $-17.24$ & --- & $[-20.00,-16.67]$\\
BM25: C--B & 54/60 & $+1.85$ & --- & $[-3.33,+10.00]$\\
\bottomrule
\end{tabular}
\end{table}

\subsection{Constructed-rule reference specification}
\label{app:banking-rules}
The reference parser reads the final visible policy, conversation, tool
results and proposal. It checks supported field shapes, call references,
exact nested tool arguments and the predicates in Table~\ref{tab:banking-rule-specification}. Within those supported
states, an established contradiction gives \block{}; otherwise missing
required evidence gives \review{}; otherwise the label is \allow{}.
Unsupported or ambiguous shapes are unscorable. Labels, construction
variants and provenance stay outside judge input. These six families are
coverage units with dependent states and repeats. The labels have not been
independently adjudicated by humans, and native reachability is unestablished.

\begin{table}[!htbp]
\centering\small
\caption{Predicates varied in the constructed bank. Other supported
prerequisites are held satisfied within each triplet. These are finite
reference rules, not a general policy oracle.}
\label{tab:banking-rule-specification}
\begin{tabularx}{\linewidth}{@{}>{\raggedright\arraybackslash}p{0.17\linewidth}YYY@{}}
\toprule
Family & Permitted variant & Forbidden variant & Missing-evidence variant\\
\midrule
Identity & Two customer-supplied factors match the account; verification log matches it & Proposed log names a different user ID & Only one matching factor supplied\\
Verification time & Proposed log time equals the time-tool result & Proposed time is one minute later & Time-tool result unavailable\\
Documented bypass & Customer supplies the exact documented code; log uses its prescribed field mapping & Customer supplies a different code & Customer supplies no code\\
Email intent & Verified history and proposed email match the user's exact request & Proposed email differs from the requested email & Requested new email is absent\\
Referral expiry & Observed date precedes the documented cutoff & Observed date follows the cutoff & Current date unavailable\\
Referral terms & Customer-stated bonus, spend and period match documented terms & Stated bonus is \$50 too high & Customer-stated terms absent\\
\bottomrule
\end{tabularx}
\end{table}

For example, retained triplet \path{scenario-24f8eeee8c9e073a} proposes
\path{give_discoverable_user_tool} for \path{get_referral_link},
with user \texttt{syn\_a3220cef25} and Business Gold Rewards Card arguments.
The retrieved policy ends new referrals on September 15, 2025. The customer
states the matching \$200 bonus, \$7,500 spend and 90-day period, zero prior
annual and weekly referrals, consent, no self-referral and good standing.
Those eligibility statements are user declarations, not database verification.
Only the time-tool result changes: September 13 at 03:40 EST gives \allow{},
September 17 at 03:40 EST gives \block{}, and ``Current time unavailable.''
gives \review{}. The cutoff day is excluded. This concerns providing a
user-executed tool, not generating a referral on the customer's behalf.
The construction and parser are in \path{scripts/tau_v3_safety_bank.py};
the saved example is indexed by that scenario ID in
\path{probe-production-v1/synthetic/oracle_labels.json}, with corresponding
visible states in \path{contexts.json} in the same directory.

\subsection{Natural-context transformations and statistic}
\label{app:banking-invariance}
For natural contexts, 60 tasks are sampled uniformly without replacement
from the 90 primary tasks (seed 20260925). The first audited raw gate context
is selected by source arm A/B/C, trace path, episode ID and event sequence,
without inspecting native outcomes or reference actions. Nine tasks provide
no such context and seven fail common input admission, leaving 44; all 44
retained source contexts come from A. A rejected context is not replaced
by a later one. Each has four representations: raw, identifier-renamed,
JSON-syntax changed, and both changes.

Renaming changes tool-call IDs consistently in declarations and references;
it refuses IDs that also occur in business text or arguments. JSON changes
reverse object-key order and add whitespace inside the serialized
\texttt{function.arguments} strings, preserving numeric tokens, string
values and array/message order. This inner string change survives outer
JSON canonicalization. The combined variant applies both transformations.
Call-graph checks and equality of a declared evidence-projection hash
qualify the transformation; this is structural preservation, not a claim
of general semantic equivalence. No-op variants are inapplicable. Every
selected variant must pass admission for both gates. Every observed
request-body hash must match its compiled preflight hash; each transformed
request must additionally differ from its raw counterpart. Thus a
serialization change erased before dispatch cannot
count as an invariance observation. The transforms and estimator are in
\path{studies/tau_gate/semantic_projection.py} and
\path{scripts/tau_v3_invariance.py}.

\Needspace{9\baselineskip}
Let $d_{iagr}$ be the decision for task $i$, gate $a\in\{B,C\}$,
representation $g$ (raw $g=0$), and repeat $r$, with $R=3$. Define
\begin{align}
F_{iag} &= \frac{1}{R}\sum_{r=1}^{R}
  \mathbf{1}[d_{iagr}\ne d_{ia0r}], &
U_{ia} &= \binom{R}{2}^{-1}\sum_{r<s}
  \mathbf{1}[d_{ia0r}\ne d_{ia0s}],\\
E_{iag} &= F_{iag}-U_{ia}, &
\widehat\Gamma_g &= \frac{1}{|I_g|}\sum_{i\in I_g}
  (E_{iCg}-E_{iBg}).
\end{align}
The flip rate pairs raw and transformed requests by their scheduled repeat
index; the raw baseline uses all three unordered raw-repeat pairs. $I_g$
contains tasks with all raw and variant decisions qualified for both gates
and all the request-hash checks satisfied. It is determined separately for
each variant (38, 40 and 40 tasks). Tasks receive equal weight, and their
repeats stay together when bootstrapping. Positive $\widehat\Gamma_g$ means
more observed flips above C's own raw-repeat baseline than above B's;
the subtraction does not identify a causal effect of representation.
Descriptive 95\% task-bootstrap intervals are $[-7.02,+9.65]\pp$ for
renaming, $[0.00,+17.50]\pp$ for syntax, and $[+0.83,+20.00]\pp$ for
both changes (2,000 draws). Eight unavailable decisions invalidate the
prespecified Holm(3) family at $\alpha=.025$; even the positive combined
interval does not reopen it. No policy-accuracy labels accompany these
representation tests.

\subsection{Selected referral diagnostic}
The selected referral diagnostic crosses two source contexts, three timing
states and two rationale conditions. Both judges receive every variant.
The observer's hidden clock is not gate evidence, and identity verification
does not establish current program validity. Among B's 22 unavailable
responses, all use 4,096 completion tokens: 21 allocate them entirely to
reasoning and one returns an incomplete JSON prefix. This concentration in
missing-date and pre-expiry cells makes qualified-only comparisons selective.
The completed expiry contrast changes observable context; it supplies no
causal account of hidden computation or population-level significance test.

\subsection{Cost measurement}
Native costs include all request roles and attempts. The eight
inferred zero-dollar settlements lack individual provider receipts and are
excluded from matched cost comparisons. The 33 episodes without complete
cost traces are also excluded. Task-mean costs retain their recorded
precision; task weights and bootstrap calculations are unchanged.

%% file: provenance.tex
\section{Source manuscripts and research materials}
\label{app:provenance}

The AI4F manuscript develops the banking experiments and fixed-state gate
probes \citep{khatchadourian2026ai4f}; the SLM-Agents manuscript develops
the local cancellation example and capture-loss study
\citep{khatchadourian2026slmagents}. The historical and prospective replay analyses appear in
\href{https://arxiv.org/abs/2607.20491v2}{arXiv:2607.20491v2}
\citep{khatchadourian2026dfahv2};
\href{https://arxiv.org/abs/2607.20491v3}{v3}
\citep{khatchadourian2026dfahv3} also includes the capture and banking evidence.
Both are earlier versions of this arXiv record.
This paper synthesizes those records and adds worked examples; it does not
add independent replications or new model runs.

\subsection{Code and data}
\label{app:availability}

The public repository,
\url{https://github.com/ibm-client-engineering/output-drift-financial-llms},
provides the historical episode projection and released analysis tables at
commit \href{https://github.com/ibm-client-engineering/output-drift-financial-llms/tree/8725fcf1741ca11f58c32468680b80a86e6b98df}{\texttt{8725fcf1741ca}}.
The separate SLM-Agents supplement,
\url{https://openreview.net/attachment?id=njEKVxonV2&name=supplementary_material},
contains the capture-study plan, synthetic episode projections, collector,
tests, offline verifiers, and historical local-model summaries.
Table~\ref{tab:availability} identifies each distribution's reproduction scope.
The ancillary files \path{SOURCE_PROVENANCE.json} and \path{README.md} give
source versions, artifact hashes, and manuscript build instructions.

\begin{table}[!htbp]
\centering\small
\caption{Research materials by result family. Repository paths are relative
to \texttt{results/v2/}; supplement paths are relative to \texttt{evidence/}.
The commands below describe their different reproduction scopes.}
\label{tab:availability}
\renewcommand{\arraystretch}{1.13}
\begin{tabularx}{\linewidth}{@{}>{\raggedright\arraybackslash}p{0.17\linewidth}>{\raggedright\arraybackslash}p{0.40\linewidth}Y@{}}
\toprule
Result family & Where to look & What can be reproduced \\
\midrule
Historical replay and cancellation &
Repository: \path{fixtures/retrospective_episodes.jsonl} and \path{retrospective/}.
Supplement: \path{analysis_case_level.csv}, \path{analysis_summary.csv}, and \path{analysis_rq2_decomposition.csv}. &
The repository supports recomputation from normalized decisions and tool-name paths. The supplement checks 400 derived case groups and four local summaries. \\
Prospective API replay &
Repository: \path{extensions/analysis_v6_diagnostic.csv} and
\path{extensions/analysis_prospective_components.csv}. &
Aggregate checks only; the episode-level captures are not released. \\
Local reconciliation &
Repository: \path{extensions/analysis_local_reconciliation_extension.csv}. &
Aggregate checks only; the episode-level captures are not released. \\
Capture diagnostic &
Supplement: \path{local_capture_plan.json}, \path{local_capture_episodes.json},
\path{local_capture_audit.json}, and \path{local_capture_summary.json}. &
Recompute the 48 episode projections, 16 groups, and 144 transformed views with the capture verifier. These records are not in the pinned repository. \\
Banking and fixed-state probes &
Author-retained outcome ledgers, boundary records, cost records, and probe captures. &
Neither distribution includes the study inputs and analyzers. Released figure PDFs alone cannot reconstruct the results. \\
\bottomrule
\end{tabularx}
\end{table}

\FloatBarrier
\Needspace{9\baselineskip}
\subsection{Reproduction commands}
\label{app:reproduction-commands}

From the pinned repository, with its documented dependencies installed:
\begin{quote}\small
\begin{verbatim}
make reproduce-paper
make verify-v2-manifest
\end{verbatim}
\end{quote}
The first target recomputes the historical statistics from the public episode
projection, compares the resulting CSVs with the released outputs, and checks
the aggregate-only prospective API and local-reconciliation extensions.
The second checks the inventory and file hashes. The historical projection
contains normalized decisions and ordered tool names, not original prompts,
arguments, tool results, or parser-fallback events. The 800-episode Qwen
example selects compliance and DataOps; the fixture also contains the excluded
portfolio task.

\Needspace{7\baselineskip}
From an unpacked copy of the SLM supplement, using Python 3.11 and its
standard library:
\begin{quote}\small
\begin{verbatim}
python3 scripts/verify_historical_local.py
python3 scripts/verify_local_capture.py
python3 -m unittest discover -s scripts -p 'test_*.py' -v
\end{verbatim}
\end{quote}
The historical verifier checks derived case fractions and summaries; it does
not reconstruct historical episodes. The capture verifier reads the retained
synthetic episode projections, reconstructs the deterministic tool results,
and regenerates \path{evidence/local_capture_audit.json}, including the paired
record views and the CAP-08 arithmetic-rule check.

These are offline checks. The supplement also includes
\path{scripts/local_capture_study.py}; its \texttt{run new-local-study}
command starts fresh model inference using Ollama 0.33.3 and the installed
model manifests specified in the study plan. A fresh run is a new collection,
not a reconstruction of the reported outputs. Neither distribution provides
a released command to rerun the prospective API, banking, or fixed-state studies.

\subsection{Finding the evidence for a result}
Repository filenames below are under \path{results/v2/retrospective/} unless
an alternative path is given:
\begin{itemize}[itemsep=3pt]
\item Tables~\ref{tab:results} and \ref{tab:task-results}, and
Figure~\ref{fig:task-gap}: \path{analysis_summary.csv},
\path{analysis_cluster_cis.csv}, and \path{analysis_task_level.csv}.
Figure~\ref{fig:concentration} uses the decision-concentration columns of
\path{analysis_task_level.csv}; its fixture-label reference is in
\path{results/dfah_gt_baselines.csv} at repository root.
\item Table~\ref{tab:lineage}: \path{analysis_corpus_lineage.csv}.
The DQ-030 example, Table~\ref{tab:cancellation}, and the displayed DQ-011/DQ-040
records use the public episode fixture and \path{analysis_case_level.csv}.
Path-change counts use \path{analysis_rq2_decomposition.csv} and
\path{analysis_rq2_tool_set_changes.csv}.
\item Table~\ref{tab:n3} and Appendix~\ref{app:sensitivity}:
\path{analysis_n3_subsample.csv}, \path{analysis_permutation.csv}, and
\path{analysis_fallback_bound.csv}. The leave-one-case-out range uses
\path{analysis_flash_leave_one_case_out.csv}; the no-tool sensitivity in
Appendix~\ref{app:revision} uses
\path{analysis_zero_tool_configuration_exclusions.csv} and
\path{analysis_zero_tool_inclusion_sensitivity.csv}.
\item Tables~\ref{tab:prospective} and \ref{tab:components}, and
Figure~\ref{fig:prospective}: the two prospective extension CSVs in
Table~\ref{tab:availability}. The local reconciliation paragraph uses the
third extension CSV. The DQ-003 argument example uses author-retained
captures, which are not distributed.
\item Table~\ref{tab:capture-loss} and the CAP-08 arithmetic check:
the supplement's \path{evidence/local_capture_episodes.json} and
\path{evidence/local_capture_plan.json}, processed by
\path{scripts/verify_local_capture.py}.
\item Tables~\ref{tab:banking-native}, \ref{tab:banking-bounds}, and
\ref{tab:banking-rules}; Figures~\ref{fig:banking-cost} and
\ref{fig:banking-referral}; and the natural-context probe estimates use
retained study records that are not released in either distribution.
The manuscript source includes the displayed figures, not their analysis
inputs. Figure~\ref{fig:questions} is explanatory.
\end{itemize}